\documentclass[10pt,twocolumn,letterpaper]{article}
\usepackage{subfloat}
\usepackage{titling}
\usepackage{iccv}
\usepackage{soul}
\usepackage{array}
\usepackage{times}
\usepackage{epsfig}
\usepackage{graphicx}
\usepackage{multirow}
\usepackage{amsmath}
\usepackage{amssymb}
\usepackage{booktabs,siunitx}
\usepackage{adjustbox}
\usepackage{algorithm}
\usepackage{algpseudocode}
\usepackage{subcaption}
\usepackage{footmisc}

\soulregister\cite7
\soulregister\ref7

\usepackage[pagebackref=true,breaklinks=true,letterpaper=true,colorlinks,bookmarks=false]{hyperref}

\hypersetup{
  plainpages=false,
  colorlinks=true,   			%
  linkcolor=blue,    			%
  anchorcolor=blue,  			%
  citecolor=blue,    			%
  filecolor=blue,    			%
  pagecolor=blue,    			%
  urlcolor=blue,     			%
  pdfview=FitH,                 %
  pdfstartview=FitH,            %
  pdfpagelayout=SinglePage      %
}

\DeclareMathOperator*{\argmin}{arg\,min}
\iccvfinalcopy

\ificcvfinal\fi

\begin{document}

\title{Joint Discovery of Object States and Manipulation Actions}
\author{
Jean-Baptiste Alayrac\thanks{D\'{e}partement d’informatique de l'ENS, \'{E}cole normale sup\'{e}rieure, CNRS, PSL Research University, 75005
	Paris, France.} \ \thanks{INRIA}
\and
Josef Sivic\footnotemark[1] \ \footnotemark[2] \ \thanks{Czech Institute of Informatics, Robotics and Cybernetics at the
Czech Technical University in Prague.}
\and
Ivan Laptev\footnotemark[1] \ \footnotemark[2]
\and
Simon Lacoste-Julien\thanks{Department of CS \& OR (DIRO), Universit\'{e} de Montr\'{e}al, Montr\'{e}al.}
}
\maketitle
\begin{abstract}
Many human activities involve object manipulations aiming to modify the object state.
Examples of common state changes include full/empty bottle, open/closed door, and attached/detached car wheel.
In this work, we seek to automatically discover the states of objects and the associated manipulation actions.
Given a set of videos for a particular task, we propose a joint model that learns to identify object states and to localize state-modifying actions.
Our model is formulated as a discriminative clustering cost with constraints.
We assume a consistent temporal order for the changes in object states and manipulation actions, 
and introduce new optimization techniques to learn model parameters without additional supervision.
We demonstrate successful discovery of seven manipulation actions and corresponding object states on a new dataset of videos depicting real-life object manipulations.
We show that our joint formulation results in an improvement of object state discovery by action recognition and vice versa.
\end{abstract}
\vspace*{-3mm}
\section{Introduction}
\label{sec:intro}

Many of our activities involve changes in object states.
We need to open a book to read it, to cut bread before eating it and to lighten candles before taking out a birthday cake.
Transitions of object states are often coupled with particular manipulation actions (open, cut, lighten).
Moreover, the success of an action is often signified by reaching the desired state of an object (whipped cream, ironed shirt) and avoiding other states (burned shirt).
Recognizing object states and manipulation actions is, hence, expected to become a key component of future systems such as wearable automatic assistants or home robots helping people in their daily tasks.

Human visual system can easily distinguish different states of objects, such as open/closed bottle or full/empty coffee cup~\cite{Brady06}.
Automatic recognition of object states and state changes, however, presents challenges as it requires distinguishing subtle changes in object appearance such as the presence of a cap on the bottle or screws on the car tire. %
Despite much work on object recognition and localization, recognition of object states has received only limited attention in computer vision~\cite{Isola15State}.

\begin{figure}
    \centering
    \includegraphics[width=\linewidth]{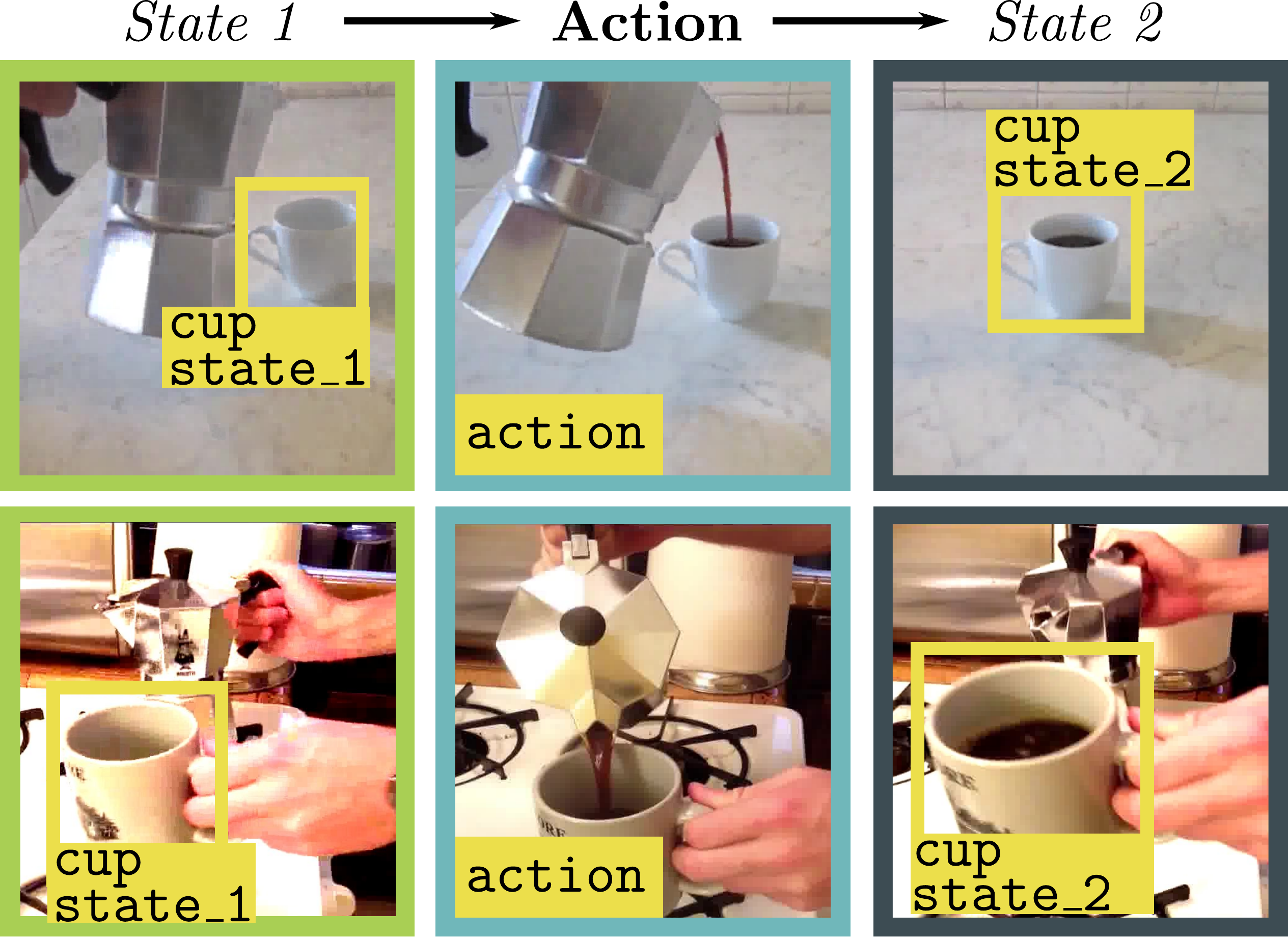}
    \caption{We automatically discover object states such as empty/full coffee cup along with their corresponding manipulation actions by observing people interacting with the objects.}
    \label{fig:teaser}
    \vspace*{-5mm}
\end{figure}

One solution to recognizing object states would be to manually annotate states for different objects,
and treat the problem as a supervised fine-grained object classification task~\cite{duan2012discovering,Farhadi09ObjectsAttrib}. This approach, however, presents two problems. First, we would have to decide {\em a priori} on the set of state labels for each object, which can be ambiguous and not suitable for future tasks. Second, for each label we would need to collect a large number of examples, which can be very costly.

In this paper we propose to {\em discover} object states directly from videos with object manipulations.
As state changes are often caused by specific actions, we attempt to jointly discover object states and corresponding manipulations.
In our setup we assume that two distinct object states are temporally separated by a manipulation action.
For example, the empty and full states of a coffee cup are separated by the ``pouring coffee" action, as shown in Figure~\ref{fig:teaser}.  
Equipped with this constraint, %
we develop a clustering approach that jointly (i)~groups object states with similar appearance and consistent temporal locations with respect to the action and (ii) finds similar manipulation actions separating those object states in the input videos.
Our approach exploits the complementarity of both subproblems and finds a joint solution for states and actions.
We formulate our problem by adopting a discriminative clustering loss~\cite{Bach07diffrac} and a joint consistency cost between states and actions.  
We introduce an effective optimization solution in order to handle the resulting non-convex loss function and the set of spatial-temporal constraints.
To evaluate our method, we collect a new video dataset depicting real-life object manipulation actions in realistic videos.
Given this dataset for training, our method demonstrates successful discovery of object states and manipulation actions.
We also demonstrate that our joint formulation gives an improvement of object state discovery by action recognition and vice versa.

\section{Related work}
\label{sec:rel_work}

Below we review related work on person-object interaction, recognizing object states, action recognition and discriminative clustering that we employ in our model. 

\noindent\textbf{Person-object interactions.}
Many daily activities involve person-object interactions. %
Modeling co-occurrences of objects and actions have shown benefits for recognizing actions in~\cite{delaitre11personaction,gupta2009observing,kjellstrom2011visual,pirsiavash2012detecting,yao2011human}.
Recent work has also focused on building realistic datasets with people manipulating objects, \eg~in instructional videos~\cite{Alayrac15Unsupervised,Malmaud15what,Sener15unsupervised} or while performing daily activities~\cite{varol16hollywood}.
We build on this work but focus on joint modeling and recognition of actions and {\em object states}.

\noindent\textbf{States of objects.}
Prior work has addressed recognition of object attributes~\cite{Farhadi09ObjectsAttrib,Parikh2011Relattrib,patterson2014sun}, which can be seen as different object states in some cases. 
Differently from our approach, these works typically focus on classifying still images, do not consider human actions and assume an {\em a priori} known list of possible attributes. %
Closer to our setting, Isola~\etal\cite{Isola15State} discover object states and transformations between them by analyzing large collections of still images downloaded from the Internet. 
In contrast, our method does not require annotations of object states. %
Instead, we use the dynamics of consistent manipulations to discover object states in the video with minimal supervision. %
In~\cite{dima2014youdo}, the authors use consistent manipulations to discover task relevant objects.
However, they do not consider object states, rely mostly on first person cues (such as gaze) and take advantage of the fact that videos are taken in a single controlled environment.

\noindent\textbf{Action recognition.}
Most of the prior work on action recognition has focused on designing features to describe time intervals of a video using motion and appearance~\cite{laptev08learning,simonyan2014two,tran2015learning,Wang13action}. This is effective for actions  such as \textit{dancing} or \textit{jumping},
however, many of our daily activities are best distinguishable by their effect on the environment.
For example, \textit{opening door}  and \textit{closing door} can look very similar using only motion and appearance descriptors but their outcome is completely different.
This observation has been used to design action models in~\cite{fathi11modeling,fernando15vidDarwin,Wang16Transformation}.
In~\cite{Wang16Transformation}, for example, the authors propose to learn an embedding in which a given action acts as a transformation of  features of the video.
In our work we localize objects and recognize  changes of their states using manipulation actions as a supervisory signal.
Related to ours is also the work of Fathi~\etal~\cite{fathi11modeling} who represent actions in egocentric videos by changes of appearance of objects (also called object states), however, their method requires manually annotated precise temporal localization of actions in training videos. 
In contrast, we focus on (non-egocentric) Internet videos depicting real-life object manipulations where actions are performed by different people in a variety of challenging indoor/outdoor environments. In addition, our model jointly learns to recognize both actions and object states with only minimal supervision.

\noindent\textbf{Discriminative clustering.}
Our model builds on unsupervised discriminative clustering methods~\cite{Bach07diffrac,singh12discPat,Xu2004maximum} that group data samples according to a simultaneously learned classifier. 
Such methods can incorporate (weak) supervision that helps to steer the clustering towards a preferred solution~\cite{Bojanowski13finding,doersch2012what,feifei2016connectionist,jain2013representing,tang14coloc}.
In particular, we build on the discriminative clustering approach of~\cite{Bach07diffrac} that has been shown to perform well in a variety of computer vision problems~\cite{Bojanowski13finding}. %
It leads to a quadratic optimization problem where different forms of supervision can be  incorporated in the form of (typically) linear constraints.
Building on this formalism, we develop a model that jointly finds object states and temporal locations of actions in the video.
Part of our object state model is related to~\cite{Joulin14efficient}, while our action model is related to~\cite{Bojanowski15weakly}.
However, we introduce new spatial-temporal constraints together with a novel joint cost function linking object states and actions, as well as new effective optimization techniques.

\noindent\textbf{Contributions.}
\label{sec:contrib}
The contributions of this work are three-fold. 
First, we develop a new discriminative clustering model that jointly discovers object states and temporally localizes associated manipulation actions in video. 
Second, we introduce an effective optimization algorithm to handle the resulting non-convex constrained optimization problem. 
Finally, we experimentally demonstrate that our model discovers key object states and manipulation actions from input videos with minimal supervision.

\begin{figure*}[t!]
   \centering
     \includegraphics[width=\linewidth]{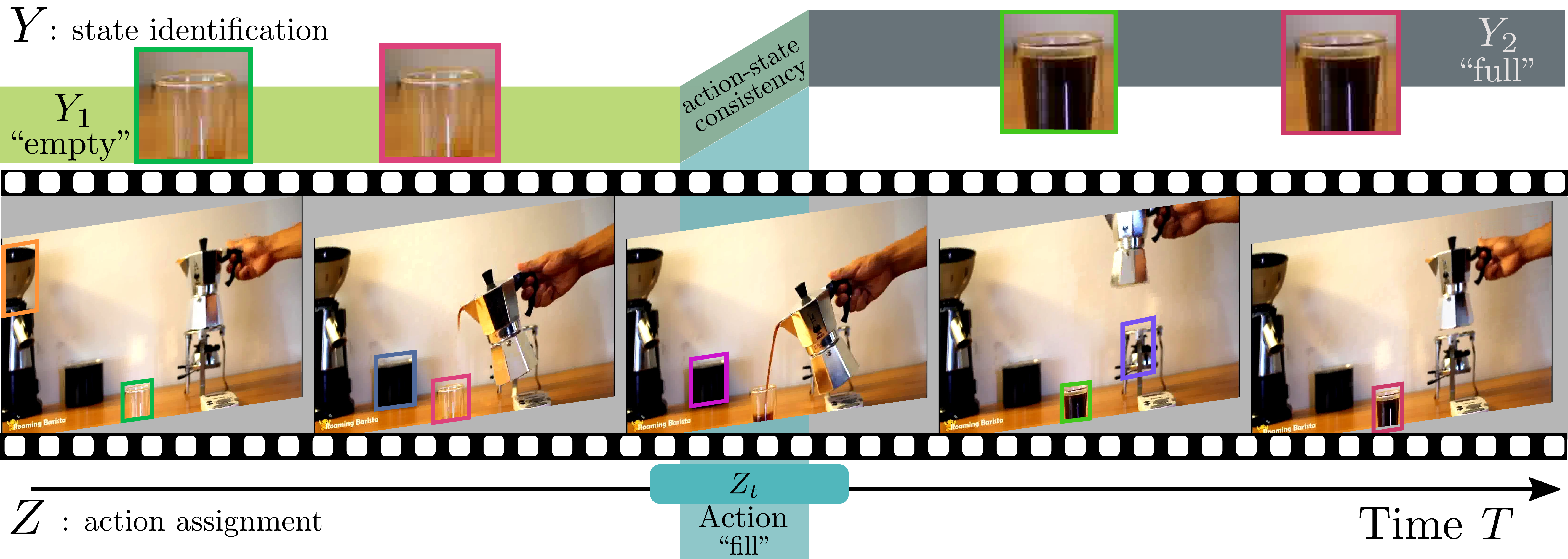}
     \caption{
     Given a set of clips that depict a manipulated object, we wish to automatically discover the main states that the object can take along with localizing the associated manipulation action.
     In this example, we show one video of someone filling a coffee cup.
     The video starts with an empty cup (\textit{state 1}), which is filled with coffee (\textbf{action}) to become full (\textit{state 2}).
     Given imperfect object detectors, we wish to assign to the valid object candidates either the initial state or the final state (encoded in~$Y$).
	 We also want to localize the manipulating action in time (encoded in~$Z$) while maintaining a joint action-state consistency.
     }
     \vspace{-2mm} %
     \label{fig:mainpaper}
 \end{figure*}	

\section{Modeling manipulated objects}
\label{sec:approach}

We are given a set of $N$ clips that contain a common \textbf{manipulation} of the same \textit{object} (such as ``\textbf{open} an \textit{oyster}").
We also assume that we are given an \emph{a priori} model of the corresponding object in the form of a pre-trained object detector~\cite{girsh15fastrcnn}.
Given these inputs, our goal is twofold: (i)~localize the temporal extent of the action and (ii)~spatially/temporally localize the manipulated object and identify its states over time.
This is achieved by jointly clustering the appearances of an object (such as an ``oyster") appearing in all clips into two classes, corresponding to the two different states (such as ``closed" and ``open"), while at the same time temporally localizing a consistent ``opening" action that separates the two states consistently in all clips.   
More formally, we formulate the problem as a minimization of a joint cost function that ties together the action prediction in time, encoded in the assignment variable~$Z$, with the object state discovery in space and time, defined by the assignment variable~$Y$:

\begin{align}
\label{eq:jointstateactionprob}
\underset{\substack{Y\in\{0,1\}^{M\times 2}\\Z\in\{0,1\}^T}}{\text{minimize}} \!\! & & & \phantom{aai} f(Z) \; +  \; g(Y) \; + \; d(Z, Y) \\[-5mm]
& & \text{ s.t. }   & \underbrace{Z \in \mathcal{Z}}_{\substack{\text{saliency of action} \\ \\ \text{\textbf{Action localization}}}} \ \text{ and }  \  \underbrace{Y \in \mathcal{Y}}_{\substack{\text{ordering + non overlap}\\ \\ \text{\textbf{Object state labeling}}}}   \nonumber
\end{align}
where $f(Z)$ is a discriminative clustering cost to temporally localize the action in each clip, $g(Y)$ is a discriminative clustering cost to identify and localize the different object states and $d(Z,Y)$ is a joint cost that relates object states and actions together. $T$ denotes the total length of all video clips and $M$ denotes the total number of tracked object candidate boxes (tracklets).
In addition, we impose constraints~$\mathcal{Y}$ and~$\mathcal{Z}$ that encode additional structure of the problem: we localize the action with its most salient time interval per clip (``saliency"); we assume that the ordering of object states is consistent in all clips (``ordering") and that only one object is manipulated at a time (``non overlap").

In the following, we proceed with describing different parts of the model~\eqref{eq:jointstateactionprob}.
In Sec.~\ref{sec:statemodel} we describe the cost function for the discovery of object states. In Sec.~\ref{sec:actionmodel} we detail our model for action localization. Finally,  in Sec.~\ref{sec:joint} we describe and motivate the joint cost $d$.
\subsection{Discovering object states}
\label{sec:statemodel}

The goal here is to both (i) spatially localize the manipulated object and (ii) temporally identify its individual states.
To address the first goal, we employ pre-trained object detectors. To address the second goal, we formulate the discovery of object states as a discriminative clustering task with constraints.
We obtain candidate object detections using standard object detectors pre-trained on large scale existing datasets such as ImageNet~\cite{imagenet09}. We assume that each clip $n$ is accompanied with a set of $M_n$ tracklets\footnote{In this work, we use short tracks of objects (less than one second) that we call tracklet. We want to avoid long tracks that continue across a state change of objects. By using the finer granularity of tracklets, our model has the ability to correct for detection mistakes within a track as well as identify more precisely the state change. 
} of the object of interest. %

We formalize the task of localizing the states of objects as a discriminative clustering problem where the goal is to find an assignment matrix~$Y_n\in\{0,1\}^{M_n\times 2}$, where
 $(Y_n)_{mk}=1$ indicates that the $m$-th tracklet represents the object in state $k$. 
 We also allow a complete row of~$Y_n$ to be zero to encode that no state was assigned to the corresponding tracklet. 
 This is to model the possibility of false positive detections of an object, or that another object of the same class appears in the video, but is not manipulated and thus is not undergoing any state change. 
 In detail, we minimize the following discriminative clustering cost~\cite{Bach07diffrac}:\footnote{We concatenate all the variables~$Y_n$ into one $M\!\times\!2$ matrix $Y\!$.}  
\begin{align}
    g(Y) &= \min_{W_{s} \in \mathbb{R}^{d_s \times 2}} \ \underbrace{\frac{1}{2M} \|Y - X_{s} W_{s}\|_F^2}_\text{Discriminative loss on data} + \underbrace{\frac{\mu}{2} \|W_{s}\|_F^2}_\text{Regularizer}
    \label{eq:discr_state} 
\end{align}
where $W_s$ is the object state classifier that we seek to learn, $\mu$ is a regularization parameter and $X_s$ is a $M\!\times\!d_s$ matrix of features, where each row is a $d_s$-dimensional (state) feature vector storing features for one particular tracklet. 
The minimization in $W_s$ actually leads to a convex quadratic cost function in~$Y$ (see~\cite{Bach07diffrac}). 
The first term in~\eqref{eq:discr_state} is the discriminative loss on the data that measures how easily the input data~$X_{s}$ is classified by the linear classifier~$W_{s}$ when the object state assignment is given by matrix~$Y$. 
In other words, we wish to find a labeling~$Y$ for given object tracklets into two states (or no state) so that their appearance features~$X$ are easily classified by a linear classifier. To steer the cost towards the right solution, we employ the following constraints (encoded by $Y \in \mathcal{Y}$ in \eqref{eq:jointstateactionprob}). 

\noindent\textbf{Only one object is manipulated at a time : non overlap constraint.}
As it is common in instructional videos, we assume that only one object can be manipulated at a given time.
However, in practice, it is common to have multiple (spatially diverse) tracklets that occur at the same time, for example, due to a false positive detection in the same frame.
To overcome this issue, we impose that at most one tracklet can be labeled as belonging to \emph{state 1} or \emph{state 2} at any given time.
We refer to this constraint as ``\emph{non overlap}" in problem~\eqref{eq:jointstateactionprob}.

\noindent\textbf{\emph{state 1}  $\rightarrow$ \emph{Action} $\rightarrow$  \emph{state 2}: ordering constraints.}
We assume that the manipulating action transforms the object from an initial state to a final state and that both states are present in each video.
This naturally introduces two constraints.
The first one is the ordering constraints on the labeling~$Y_n$, \ie the \emph{state 1} should occur before \emph{state 2} in each video.
The second constraint imposes that we have \emph{at least one} tracklet labeled as \emph{state 1} and at least one tracklet labeled as \emph{state 2}.
We call this last constraint the ``\textbf{at least one}" constraint in contrast to forcing ``exactly one" ordered prediction as previously proposed in a discriminative clustering approach on video for action localization~\cite{Bojanowski15weakly}.
This new type of constraint brings additional optimization challenges that we address in Section~\ref{subsec:frank-wolfe}.

\subsection{Action localization}
\label{sec:actionmodel}

Our action model is equivalent to the one of~\cite{Bojanowski15weakly} applied to only \emph{one action}. 
More precisely, the goal is to find an assignment matrix~$Z_n\in\{0,1\}^{T_n}$ for each clip $n$, where
$Z_{nt}=1$  encodes that the $t$-th time interval of video is assigned to an action and $Z_{nt}=0$ encodes that no action is detected in interval $t$.
The cost that we minimize for this problem is similar to the object states cost:
\vspace{-0.6mm}
\begin{align}
	f(Z) = \min_{W_v \in \mathbb{R}^{d_v}} \ \underbrace{\frac{1}{2T} \|Z - X_{v} W_{v}\|_F^2}_\text{Discriminative loss on data} + \underbrace{\frac{\lambda}{2} \|W_{v}\|_F^2}_\text{Regularizer} ,  
	\label{eq:actioncost} 
\end{align}
where $W_v$ is the action classifier, $\lambda$ is a regularization parameter and $X_v$ is a matrix of visual features.
We constrain our model to predict \emph{exactly} one time interval for an action per clip, an approach for actions that was shown to be beneficial in a weakly supervised setting~\cite{Bojanowski15weakly} (referred to as ``\emph{action saliency}" constraint).
As will be shown in experiments, this model \emph{alone} is incomplete because the clips in our dataset can contain other actions that do not manipulate the object of interest. 
Our central contribution is to propose a \emph{joint formulation} that links this action model with the object state prediction model, thereby resolving the ambiguity of actions.
We detail the joint model next.

\subsection{Linking actions and object states}
\label{sec:joint}
Actions in our model are directly related to changes in object states.
We therefore want to enforce consistency between the two problems.
To do so, we design a novel joint cost function that operates on the action video labeling~$Z_n$ and the state tracklet assignment~$Y_n$ for each clip.
We want to impose a constraint that the action occurs in between the presence of the two different object states.
In other words, we want to penalize the fact that state~$1$ is detected $\emph{after}$ the action happens, or the fact that state~$2$ is triggered  $\emph{before}$ the action occurs.

\noindent\textbf{Joint cost definition.}
We propose the following joint symmetric cost function for each clip:
\vspace{-0.6mm}
\begin{align}
d(Z_n,Y_n) =   \hspace{-0.5em}\sum_{y\in\mathcal{S}_1(Y_n)} \!\!\! [t_y-t_{Z_n}]_{+}  + \hspace{-0.5em}\sum_{y\in\mathcal{S}_2(Y_n)}  \! [t_{Z_n} \!\!-t_y]_{+},
\label{eq:dist}
\end{align}
where $t_{Z_n}$ and $t_y$ are the times when the action~$Z_n$ and the tracklet $y$ occur in a clip~$n$, respectively.
$\mathcal{S}_1(Y_n)$ and $\mathcal{S}_2(Y_n)$ are the tracklets in the $n$-th clip that have been assigned to state $1$ and state $2$, respectively.
Finally $[x]_{+}$ is the positive part of $x$.
In other words, the function penalizes the inconsistent assignment of objects states~$Y_n$  by the
amount of time that separates the incorrectly assigned tracklet and the manipulation action in the clip.
The overall joint cost is the sum over all clips weighted by a scaling hyperparameter $\nu>0$:
\vspace{-0.7mm}
\begin{equation} 
d(Z,Y) = \nu \frac{1}{T} \sum_{n=1}^N d(Z_n,Y_n). \label{eq:distGlobal}
\end{equation}
\section{Optimization}
\label{sec:optim}

Optimizing problem~\eqref{eq:jointstateactionprob} poses several challenges that need to be addressed.
First, we propose a relaxation of the integer constraints and the distortion function (Section~\ref{sec:relax}).
Second, we optimize this relaxation using Frank-Wolfe with a new dynamic program able to handle our tracklet constraints (Section~\ref{subsec:frank-wolfe}).
Finally, we introduce a new rounding technique to obtain an integer candidate solution to our problem (Section~\ref{sec:rounding}).

\subsection{Relaxation}
\label{sec:relax}
Problem~\eqref{eq:jointstateactionprob} is NP-hard in general~\cite{loiola07qap} due to its specific integer constraints.
Inspired by the approach of~\cite{Bojanowski14weakly} that was successful to approximate combinatorial optimization problems, we propose to use the tightest convex relaxation of the feasible subset of binary matrices by taking its convex hull.
As our variables now can take values in $[0,1]$, we also have to propose a consistent extension for the different cost functions to handle fractional values as input.
For the cost functions $f$ and $g$, we can directly take their expression on the relaxed set as they are already expressed as (convex) quadratic functions.
Similarly, for the joint cost function $d$ in~\eqref{eq:dist}, we use its natural bilinear relaxation:

\vspace{-3mm}
\begin{align}
d(Z_n,Y_n) = \sum_{i=1}^{M_n} \sum_{t=1}^{T_n}  \Big( &(Y_n)_{i1} Z_{nt} [t_{ni}-t]_+ \,\, + \notag \\[-3mm]
&(Y_n)_{i2} Z_{nt} [t-t_{ni}]_+ \,\, \Big),
\label{eq:relaxeddist}
\end{align}
where $t_{ni}$ denotes the video time of tracklet $i$ in clip $n$.
This relaxation is equal to the function~\eqref{eq:dist} on the integer points.
However, it is not jointly convex in~$Y$ and~$Z$, thus we have to design an appropriate optimization technique to obtain good (relaxed) candidate solutions, as described next.

\subsection{Joint optimization using Frank-Wolfe}
\label{subsec:frank-wolfe}
When dealing with a constrained optimization problem for which it is easy to solve linear programs but difficult to project on the feasible set, the Frank-Wolfe algorithm is an excellent choice~\cite{Jaggi2013,Lacoste15GlobalLinearFW}.
It is exactly the case for our relaxed problem, where the linear program over the convex hull of feasible integer matrices can be solved efficiently via dynamic
programming.
Moreover, \cite{lacoste16nonconvexFW} recently showed that the Frank-Wolfe algorithm with line-search converges to a stationary point for non-convex objectives at a rate of $O(1/\sqrt{k})$.
We thus use this algorithm for the joint optimization of~\eqref{eq:jointstateactionprob}. As the objective is quadratic, we can perform exact line-search analytically, which speeds up convergence in practice.
Finally, in order to get a good initialization for both variables $Z$ and $Y$, we first optimize separately $f(Z)$ and $g(Y)$ (without the non-convex $d(Z,Y)$), which are both convex functions.

\noindent\textbf{Dynamic program for the tracklets.}
In order to apply the Frank-Wolfe algorithm, we need to solve a linear program (LP) over our set of constraints. 
Previous work has explored ``\textit{exact one}" ordering constraints for time localization problems~\cite{Bojanowski14weakly}.
Differently here, we have to deal with the spatial (non overlap) constraint  \emph{and} finding ``\textit{at least one}" candidate tracklet per state.
To deal with these two requirements, we propose a novel dynamic programming approach.
First, the ``at least one" constraint is encoded by having a memory variable which indicates whether state~1 or state~2 have already been visited. This variable is used to propose valid state decisions for consecutive tracklets.
Second, the challenging ``non-overlap" tracklet constraint is included by constructing valid left-to-right paths in a cost matrix while carefully considering the possible authorized transitions.  
We provide details of the formulation in Appendix~\ref{app:dp}.
In addition, we show in section~\ref{sec:exp_res} that these new constraints are \emph{key} for the success of the method. %

\subsection{Joint rounding method}
\label{sec:rounding}
Once we obtain a candidate solution of the relaxed problem, we have to round it to an integer solution in order to make predictions.
Previous works~\cite{Alayrac15Unsupervised,Bojanowski15weakly} have observed that using the learned $W^*$ classifier for rounding gave better results than other possible alternatives. We extend this approach to our joint setup by proposing the following new rounding procedure. 
We optimize problem~\eqref{eq:jointstateactionprob} but fix the values of $W$ in the discriminative clustering costs. 
Specifically, we minimize the following cost function over the integer points $Z\in \mathcal{Z}$ and $Y \in \mathcal{Y}$:
\begin{align}
\label{eq:jcrrounding}
\!\!\!\!\!\frac{1}{2T} \|Z - X_{v} W^*_{v}\|_F^2 +  \frac{1}{2M} \|Y - X_{s} W^*_{s}\|_F^2 +  d(Z, Y), \!\!
\end{align}
where $W^*_{v}$ and $W^*_{s}$ are the classifier weights obtained at the end of the relaxed optimization.
Because $y^2 = y$ when $y$ is binary, \eqref{eq:jcrrounding} is actually a \emph{linear} objective over the binary matrix~$Y_n$ for~$Z_n$ fixed.
Thus we can optimize~\eqref{eq:jcrrounding} \emph{exactly} by solving a dynamic program on $Y_n$ for each of the $T_n$ possibilities of $Z_n$, yielding $O(M_n T_n)$ time complexity per clip (see Appendix~\ref{app:jcr} for details).

\section{Experiments}
\label{sec:experiments}

In this section, we first describe our dataset, the object tracking pipeline and the feature representation for object tracklets and videos (Section~\ref{sec:dataset}). 
We consider two experimental set-ups. 
In the first weakly-supervised set-up (Section~\ref{sec:exp_res}), we apply our method on a set of video clips which we know contain the action of interest but do not know its precise temporal localization. 
In the second, more challenging ``in the wild" set-up (Section~\ref{sec:exp_weak}), the input set of weakly-supervised clips is obtained by automatic processing of text associated with the videos and hence may contain erroneous clips that do not contain the manipulation action of interest.         
The data and code are available online~\cite{Alayrac16ObjectStatesWeb}. 

\subsection{Dataset and features}
\label{sec:dataset}

\noindent\textbf{Dataset of manipulation actions.}
We build a dataset of manipulation actions by collecting videos from different sources: the instructional video dataset introduced in~\cite{Alayrac15Unsupervised}, the Charades dataset from~\cite{varol16hollywood}, and some additional videos downloaded from YouTube. We focus on ``third person" videos (rather than egocentric) as such videos depict a variety of people in different settings and can be obtained on a large scale from YouTube.
We annotate the precise temporal extent of seven different actions\footnote{\textit{put the wheel on the car (47 clips)}, \textit{withdraw the wheel from the car (46)}, \textit{place a plant inside a pot (27)}, \textit{open an oyster (28)}, \textit{open a refrigerator (234)}, \textit{close a refrigerator (191)} and \textit{pour coffee (57)}.} applied to five distinct objects\footnote{\textit{car wheel}, \textit{flower pot}, \textit{oyster}, \textit{refrigerator} and \textit{coffee cup}.}. 
This results in 630 annotated occurrences of ground truth manipulation action. 

To evaluate object state recognition, we define a list of two states for each object. %
We then run automatic object detector for each involved object, track the detected object occurrences throughout the video and then subdivide the resulting long tracks into short tracklets.
Finally, we label ground truth object states for tracklets within $\pm$40 seconds of each manipulation action.  
We label four possible states: \textit{state 1}, \textit{state 2}, \textit{ambiguous state} or \textit{false positive detection}.
The ambiguous state covers the (not so common) in-between cases, such as cup half-full.  
In total, we have 19,499 fully annotated tracklets out of which: $35\%$ cover \textit{state 1} or \textit{state 2}, $25\%$ are ambiguous, and $40\%$ are false positives.
Note that this annotation is only used for evaluation purpose, and not by any of our models.
Detailed statistics of the dataset are given in Appendix~\ref{app:dataset}.

\noindent\textbf{Object detection and tracking.}
In order to obtain detectors for the five objects, we finetune the FastRCNN network~\cite{girsh15fastrcnn} with training data from ImageNet~\cite{imagenet09}.
We use bounding box annotations from ImageNet when available (\eg the ``wheel" class).
For the other classes, we manually labeled more than 500 instances per class.
In our set-up with only moderate amount of training data, we observed that class-agnostic object proposals combined with FastRCNN performed better than FasterRCNN~\cite{ren2015faster}.
In detail, we use geodesic object proposals~\cite{kra2014gop} and set a relatively low object detection threshold ($0.4$) to have good recall.
We track objects using a generic KLT tracker from~\cite{Bojanowski13finding}.
The tracks are then post-processed into shorter tracklets that last about one second and thus are likely to have only one object state.

\noindent\textbf{Object tracklet representation.}
For each detected object, represented by a set of bounding boxes over the course of the tracklet, we compute a CNN feature from each (extended) bounding box that we then average over the length of the tracklet to get the final representation.
The CNN feature is extracted with a ROI pooling~\cite{ren2015faster} of ResNet50~\cite{he16resnet}.
The ROI pooling notably allows to capture some context around the object which is important for some cases (\eg \textit{wheel} ``on" or ``off" the car).
The resulting feature descriptor of each object tracklet is 8,192 dimensional.

\noindent\textbf{Representing video for recognizing actions.}
Following the approach of~\cite{Alayrac15Unsupervised,Bojanowski14weakly,Bojanowski15weakly}, each video is divided into chunks of 10 frames that are represented by a motion and appearance descriptor averaged over 30 frames.
For the motion we use a 2,000 dimensional bag-of-word representation of histogram of local optical flow (HOF) obtained from Improved Dense Trajectories~\cite{Wang13action}.
Following~\cite{Alayrac15Unsupervised}, we add an appearance vector that is obtained from a 1,000 dimensional bag-of-word vector of conv5 features from VGG16~\cite{Simonyan14vggnets}.
This results in a 3,000 dimensional feature vector for each chunk of 10 frames.

\begin{table*}[t!]
	\centering
	\resizebox{0.74\textwidth}{!}{
		\begin{tabular}{clccccccc|c}
			\noalign{\hrule height 1.3pt}
			\multicolumn{1}{l}{}                                                                                     & \multicolumn{1}{c}{}                         & put           & remove        & fill          & open          & fill          & open          & close         & \multicolumn{1}{l}{}                          \\
			\multicolumn{1}{l}{}                                                                                     & \multicolumn{1}{c}{\multirow{-2}{*}{Method}} & wheel         & wheel         & pot           & oyster        & coff.cup      & fridge        & fridge        & \multicolumn{1}{l}{\multirow{-2}{*}{\textbf{Average}}} \\ \noalign{\hrule height 1pt}
			
			& (\textbf{a}) Chance                        & 0.10         & 0.11          & 0.10          & 0.07          & 0.06 & 0.10         & 0.10          & 0.09                                          \\
			
			& (\textbf{b}) Kmeans                                   & 0.25          & 0.12          & 0.11          & 0.23          & 0.14          & 0.19          & 0.22          & 0.18                                         \\
			& (\textbf{c}) Constraints only                       & 0.35          & 0.38          & 0.35          & 0.36          & 0.31 & 0.29         & 0.42          & 0.35                                          \\
			
			& (\textbf{d}) Salient state only                       & 0.35          & 0.48          & 0.35          & 0.38          & 0.30          & 0.40          & 0.37          & 0.38                                          \\
			
			& (\textbf{e}) At least one state only                  & 0.43 & 0.55 & 0.46          & 0.52          & 0.29          & 0.43          & 0.39          & 0.44                                          \\

			& (\textbf{f}) Joint model                              & \textbf{0.52}          & 0.59          & \textbf{0.50} & 0.45 & 0.39          & \textbf{0.47} & \textbf{0.47} & 0.48                                 \\
			
			&(\textbf{g}) Joint model + det. scores.                                             & 0.47          & \textbf{0.65}          & \textbf{0.50}          & \textbf{0.61}        & \textbf{0.44}          & 0.46          & 0.43          & \textbf{0.51}                                         \\ \cline{2-10} 
			
			\multirow{-8}{*}{\textbf{\begin{tabular}[c]{@{}c@{}}State\\ discovery\end{tabular}}} & (\textbf{h}) Joint + GT act. feat.                    & 0.55          & 0.56          & 0.56          & 0.52          & 0.46          & 0.45          & 0.49          & 0.51                                          \\ \noalign{\hrule height 1pt}
			& (\textbf{i})\phantom{ii} Chance                                   & 0.31          & 0.20          & 0.15          & 0.11          & 0.40          & 0.23          & 0.17          & 0.22                                          \\
			& (\textbf{ii})\phantom{i} \cite{Bojanowski15weakly}      & 0.24          & 0.13          & 0.11          & 0.14          & 0.26          & 0.29          & 0.23          & 0.20                                          \\
			& (\textbf{iii}) \cite{Bojanowski15weakly} + object cues                          & 0.24          & 0.13          & 0.26          & 0.07          & \textbf{0.84} & 0.33          & 0.37          & 0.32                                          \\
			& (\textbf{iv})\phantom{i} Joint model                             & \textbf{0.67} & \textbf{0.57} & \textbf{0.48} & \textbf{0.32} & 0.82          & \textbf{0.57} & \textbf{0.44} & \textbf{0.55}                                 \\ \cline{2-10} 
			\multirow{-5}{*}{\textbf{\begin{tabular}[c]{@{}c@{}}Action \\ localization\end{tabular}}}                      & (\textbf{v})\phantom{i\textbf{i}} Joint + GT stat. feat.                   & 0.72          & 0.66          & 0.44          & 0.46          & 0.86          & 0.55          & 0.44          & 0.59                                          \\ \noalign{\hrule height 1.3pt}
		\end{tabular}
	}
	\vspace{-2mm}
	\caption{\small State discovery (top) and action localization results (bottom). \label{tab:quantres}}
\vspace*{-3mm}	
\end{table*}

\subsection{Weakly supervised object state discovery}
\label{sec:exp_res}

\noindent\textbf{Experimental setup.}
We first apply our method in a weakly supervised set-up where for each action we provide an input set of clips, where we know the action occurs somewhere in the clip
but we do not provide the precise temporal localization.  Each clip may contain other actions that affect other objects or actions that do not affect any object at all (e.g. walking / jumping). The input clips are about 20s long and are obtained by taking approximately $\pm$ 10s of each annotated manipulation action.  

\noindent\textbf{Evaluation metric: average precision.}
For all variants of our method, we use the rounded solution that reached the smallest objective during optimization.
We evaluate these predictions with a precision score averaged over all the videos.
A temporal action localization is said to be correct if it falls within the ground truth time interval.
Similarly, a state prediction is correct if it matches the ground truth state.\footnote{In particular, we count ``\textit{ambiguous}" labels as incorrect.}
Note that a ``precision" metric is reasonable in our set-up as our method is forced to predict in all videos, i.e. the recall level is fixed to all videos and the method cannot produce high precision with low recall. 

\noindent\textbf{Hyperparameters.}
In all methods that involve a discriminative clustering objective, we used $\lambda=10^{-2}$ (action localization) and $\mu=10^{-4}$ (state discovery) for all 7 actions. 
For joint methods that optimize~\eqref{eq:jointstateactionprob}, we set the weight $\nu$ of the distortion measure~\eqref{eq:distGlobal} to $1$.

\noindent\textbf{State discovery results.}
Results are shown in the top part of Table~\ref{tab:quantres}.
In the following, we refer to ``State only" whenever we use our method without looking at the action cost or the distortion measure~\eqref{eq:jointstateactionprob}.
We compare to two baselines for the state discovery task.
Baseline~(\textbf{a}) evaluates chance performance.
Baseline~(\textbf{b}) performs K-means clustering of the tracklets with $K=3$ (2 clusters for the states and 1 for false positives).
We report performance of the best assignment for the solution with the lowest objective after 10 different initializations.
Baseline~(\textbf{c}) is obtained by running our ``State only" method while using random features for tracklet representation as well as "at least one ordering" and "non overlap" constraints. We use random features to avoid non-trivial analytic derivation for the "Constraints only" performance. This baseline reveals the difficulty of the problem and quantifies improvement brought by the ordering constraints.
The next two methods are ``State only" variants.
Method~(\textbf{d}) corresponds to a replacement of the ``at least one constraint" by an ``exactly one constraint" while the method~(\textbf{e}) uses our new constraint.
Finally, we report three joint methods that use our new joint rounding technique~\eqref{eq:jcrrounding} for prediction.
Method~(\textbf{f}) corresponds to our joint method that optimizes~\eqref{eq:jointstateactionprob}.
Method~(\textbf{g}) is a simple improvement taking into account object detection score in the objective (details below). %
Finally, method~(\textbf{h}) is our joint method but using the action ground truth labels as video features in order to test the effect of having perfect action localization for the task of object state discovery. 

We first note that method~(\textbf{e}) outperforms~(\textbf{d}), thus highlighting the importance of the ``at least one" constraint for modeling object states.
While the saliency approach (taking only the most confident detection per video) was useful for action modeling in~\cite{Bojanowski15weakly}, it is less suitable for our set-up where multiple tracklets can be in the same state.
The joint approach with actions~(\textbf{f}) outperforms the ``State only" method~(\textbf{e}) on 6 out of 7 actions and obtains better average performance,  confirming the benefits of joint modeling of actions and object states.
Using ground truth action locations further improves results (cf.~(\textbf{h}) against (\textbf{f})).
Our weakly supervised approach~(\textbf{f}) performs not much lower compared to using ground truth actions (\textbf{h}), except for the states of the coffee cup (empty/full). In this case we observe that a high number of false positive detections confuses our method.
A simple way to address this issue is to add the object detection score into the objective of our method, which then prefers to assign object states to higher scoring object candidates further reducing the effect of false positives.  
This can be done easily by adding a linear cost reflecting the object detection score to objective~\eqref{eq:jointstateactionprob}.
We denote this modified method ``(\textbf{g}) Joint model + det. scores". 
This method achieves the best average performance and highlights that additional information can be easily added to our model.

\noindent\textbf{Action localization results.}
We compare our method to three different baselines and give results in the bottom part of Table~\ref{tab:quantres}.
Baseline~(\textbf{i}) corresponds to chance performance, where the precision for each clip is simply the proportion of the entire clip taken by the ground truth time interval.
Baseline~(\textbf{ii}) is the method introduced in~\cite{Bojanowski15weakly} used here with only one action. It also corresponds to a special case of our method where the object state part of the objective in equation~\eqref{eq:jointstateactionprob} is turned off (salient action only).
Interestingly, this baseline is actually worse than chance for several actions. This is because without additional information about objects, this method localizes  other common actions in the clip and not the action manipulating the object of interest. This also demonstrates the difficulty of our experimental set-up where the input video clips often contain multiple different actions.
To address this issue, we also evaluate baseline~(\textbf{iii}), which complements~\cite{Bojanowski15weakly} with the additional constraint that the action prediction has to be within the first and the last frame where the object of interest is detected, improving the overall performance above chance.
Our joint approach~(\textbf{iv}) consistently outperforms these baselines on all actions, thus showing again the strong link between object states and actions.
Finally, the approach~(\textbf{v}) is the analog of method~(\textbf{g}) for action localization where we use ground truth state labels as tracklet features in our joint formulation showing that the action localization can be further improved with better object state descriptors.
In addition, we also compare to a supervised baseline.
The average obtained performance is 0.58 which is not far from our method. 
This demonstrates the potential of using object states for action localization. 
More details on this experiment are provided in Appendix~\ref{app:supervised}.

\begin{figure}
	\raggedleft    
	\includegraphics[width=0.99\linewidth]{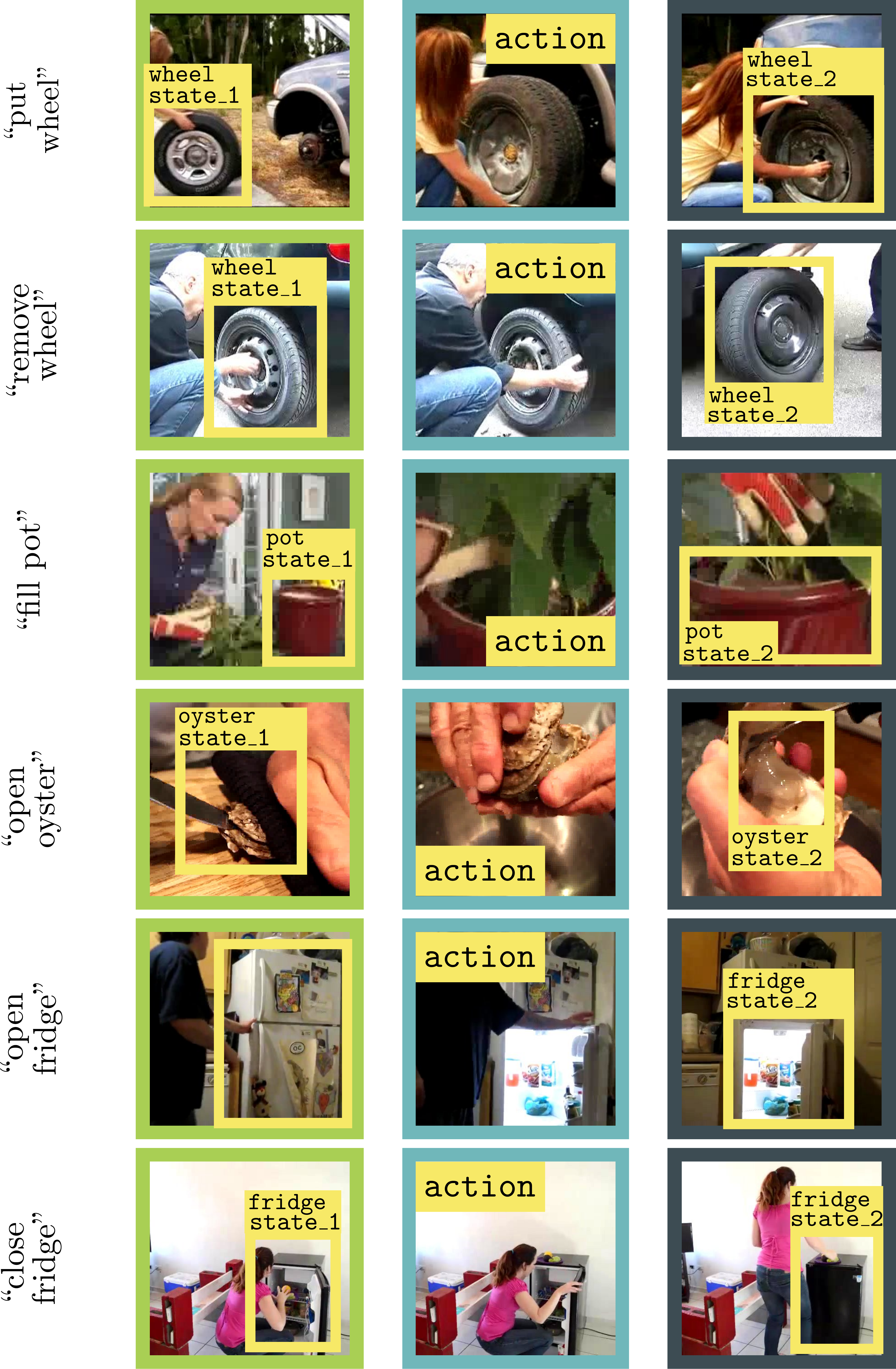}
	\caption{\small Qualitative results for joint action localization (middle) and state discovery (left and right) (see Fig.~\ref{fig:teaser} for ``\textit{fill coffee cup}").}
	\vspace*{-.5cm}
	\label{fig:qualres}
\end{figure}

\noindent\textbf{Benefits of joint object-action modeling.}
We observe that the joint modeling of object states and actions benefits both tasks.
This effect is even stronger for actions.  
Intuitively, knowing perfectly the object states reduces a lot the search space for action localization. 
Moreover, despite the recent major progress in object recognition using CNNs, action recognition still remains a hard problem with much room for improvement. 
Qualitative results are shown in Fig.~\ref{fig:qualres} and failure cases of our method are discussed in~\ref{app:failcases}.
	
\subsection{Object state discovery in the wild}
\label{sec:exp_weak}
Towards the discovery of a large number of manipulation actions and state changes, we next apply our method in an automatic setting, where action clips have been obtained using automatic text-based retrieval.

\noindent\textbf{Clip retrieval by text.}
Instructional videos~\cite{Alayrac15Unsupervised,Malmaud15what,Sener15unsupervised} usually come with a narration provided by the speaker describing the performed sequence of actions. In this experiment, we keep only such {\em narrated instructional videos} from our dataset. This results in the total of 140 videos that are 3 minutes long in average. 
We extract the narration in the form of subtitles associated with the video. 
These subtitles have been directly downloaded from YouTube and have been obtained either by Youtube's Automatic Speech Recognition (ASR) or provided by the users.%

We use the resulting text to retrieve clip candidates that may contain the action modifying the state of an object. 
Obtaining the approximate temporal location of actions from the transcribed narration is still very challenging due to ambiguities in language (``undo bolt" and ``loosen nut" refer to the same manipulation) and only coarse temporal localization of the action provided by the narration. %
Given a manipulation action such as ``remove tire", we first find positive and negative sentences relevant for the action from an instruction website such as Wikihow.
We then train a linear SVM classifier~\cite{cortes95SVM} on bigram text features. 
Finally, we use the learned classifier to score clips from the input instructional videos.
In detail, the classifier is applied in a sliding window of 10 words finding the best scoring window in each input video. The clip candidates are then obtained by trimming the input videos 5 seconds before and 15 seconds after the timing of the best scoring text window to account for the fact that people usually perform the action after having talked about it.
We apply our method on the top 20 video clips based on the SVM score for each manipulation action.
More details about this process are provided in Appendix~\ref{app:svm}.

\noindent\textbf{Results.}
As shown in Table~\ref{tab:weakres}, the pattern of results, where our joint method performs the best, is similar to the weakly supervised set-up described in Sec.~\ref{sec:exp_res}. This highlights the robustness of our model to noisy input data -- an important property for scaling-up the method to Internet scale datasets.
To assess how well our joint method could do with perfect retrieval,  
we also report results for a ``Curated" set-up where we replace the automatically retrieved clips with the 20s clips used in Sec.~\ref{sec:exp_res} for the corresponding videos. 

\begin{table}[t!]
	\centering
	\resizebox{\columnwidth}{!}{
		\begin{tabular}{@{}clccccc|r@{}}
			\noalign{\hrule height 1.3pt}
			\multicolumn{1}{l}{}                                                               & \multicolumn{1}{c}{\textbf{Method}} & \begin{tabular}[c]{@{}c@{}}put \\ wheel\end{tabular} & \begin{tabular}[c]{@{}c@{}}remove\\ wheel\end{tabular} & \begin{tabular}[c]{@{}c@{}}fill \\ pot\end{tabular} & \begin{tabular}[c]{@{}c@{}}open \\ oyster\end{tabular} & \begin{tabular}[c]{@{}c@{}}fill\\ coff.cup\end{tabular} & \multicolumn{1}{c}{\textbf{Ave.}} \\ \midrule
			\multirow{4}{*}{\textbf{\begin{tabular}[c]{@{}c@{}}State \\ disc.\end{tabular}}}   & \textbf{(c)} Cstrs only                              & 0.23                                                & 0.34                                                  & 0.25                                               & 0.29                                                   & 0.11                                                    & 0.24                              \\
			& State + det. sc.                         & 0.33                                                 & 0.48                                                  & \textbf{0.28}                                       &  0.40                                         & 0.13                                                    & 0.32                              \\
			& \textbf{(g)} Joint                               & \textbf{0.38}                                        & \textbf{0.53}                                          & 0.25                                       & \textbf{0.43 }                                                  & \textbf{0.20}                                           & \textbf{0.36}                     \\ \cmidrule(l){2-8} 
			& \textbf{(g)} Curated                             & 0.63                                                & 0.68                                                   & 0.63                                                & 0.63                                                   & 0.53                                                    & 0.62                             \\ \noalign{\hrule height 1pt}
			\multirow{4}{*}{\textbf{\begin{tabular}[c]{@{}c@{}}Action \\ local.\end{tabular}}} & \textbf{(i)} Chance                              & 0.14                                                 & 0.10                                                   & 0.06                                                & 0.10                                                   & 0.15                                                    & 0.11                             \\
			& \textbf{(iii)} Action                         & 0.05                                                & 0.10                                                   & 0.00                                                & 0.15                                          & \textbf{0.25}                                                    & 0.11                              \\
			& (\textbf{iv}) Joint                               & \textbf{0.30}                                        & \textbf{0.30}                                          & \textbf{0.20}                                       & \textbf{0.20}                                                   & 0.20                                           & \textbf{0.24}                              \\ \cmidrule(l){2-8} 
			& (\textbf{iv}) Curated                             & 0.53                                                 & 0.35                                                   & 0.32                                                & 0.40                                                   & 0.59                                                    & 0.44                              \\ \noalign{\hrule height 1.3pt}
		\end{tabular}
	}
	\vspace*{-2mm}
	\caption{\small Results on noisy clips automatically retrieved by text. \label{tab:weakres}}
	\vspace{-4mm}
\end{table}

\section{Conclusion and future work}
\vspace{-2mm}
We have described a joint model that relates object states and manipulation actions.
Given a set of input videos, our model both localizes the manipulation actions \emph{and} discovers the corresponding object states. 
We have demonstrated that our joint approach improves performance of both object state recognition and action recognition. 
More generally, our work provides evidence that actions should be modeled in the larger context of goals and effects.
Finally, our work opens up the possibility of Internet-scale learning of manipulation actions from narrated video sequences.

\footnotesize{
	\paragraph{Acknowledgments}
	This research was supported in part by a Google Research Award, ERC grants Activia (no. 307574) and LEAP (no. 336845), the CIFAR Learning in Machines \& Brains program and ESIF, OP Research, development and education Project IMPACT No. CZ.02.1.01/0.0/0.0/15 003/0000468.
}

{\small
\bibliographystyle{ieee}
\bibliography{biblio}
}

\clearpage
\appendix
\normalsize
\section*{Outline of Appendix}
This Appendix gives additional details and quantitative results for our method.
The organization is as follows.
In Appendix~\ref{app:svm}, we provide additional experimental details about the SVM training used to retrieve the clips with subtitles, described in Section~\ref{sec:exp_weak} of the main paper, along with a visualization of results.
In Appendix~\ref{app:dataset}, we give additional statistics and details about the dataset that was briefly introduced in Section~\ref{sec:dataset} of the main paper.
In Appendix~\ref{app:dp}, we give additional details about the dynamic program that we use to solve the linear program over the track constraints defined in Section~\ref{sec:statemodel} of the main paper as needed for the Frank-Wolfe optimization algorithm.
In Appendix~\ref{app:jcr}, we detail how we implement the \emph{new} joint rounding method~\eqref{eq:jcrrounding} that was introduced in Section~\ref{sec:joint} of the main paper.
In Appendix~\ref{app:supervised}, we give additional details about the supervised baselines results given in Section~\ref{sec:exp_res}.
Finally, in Appendix~\ref{app:failcases}, we comment on the common failure cases of the method.

\section{SVM training for clip retrieval}
\label{app:svm}

In Section~\ref{sec:exp_weak} we proposed an automatic method for retrieving video clips with manipulated objects.
This method makes use of narrations that come along with instructional videos. 
Narrations in the form of text are first obtained \emph{automatically} with Automatic Speech Recognition (ASR)\footnote{The ASR transcriptions are directly downloaded from YouTube.} and then processed as detailed below.

\textbf{Language dataset.}
For each manipulation action, we first find relevant positive and negative sentences on instruction websites such as Wikihow.
On average we obtain about 12 positive and 50 negative sentences per action.

\textbf{Language features.}
Off-the-shelf methods for text parsing typically fail in the absence of punctuation.
To process ASR output, which comes without punctuation, we propose to use the following simple but robust text representation. 
We represent every 10-word window of the narration by a TF-IDF vector based of uni-grams and bi-grams.
We use the same TF-IDF representation to encode text in our Language dataset on the level of sentences.

\textbf{SVM training.}
We train binary linear SVM classifiers to identify manipulation actions using the Language dataset for training and the regularization parameter $C=10$.
The obtained classifiers are then used to score every 10-word window of text narrations.
Video clips with the temporal correspondence to the top-scoring text narrations for each action are then retrieved. 
To deduce the temporal extent of the video clip given the top-scoring window, we trim 5 seconds before and 15 seconds after the corresponding timing.
This is to account for the fact that people are usually doing the action after speaking about it.
These are the clips we use for our evaluation in Section~\ref{sec:exp_weak} of the main paper.

\textbf{Illustration.} 
In Figure~\ref{tab:svm_1}, we provide an illustration of text based clip retrieval.
This visualization demonstrates the difficulty of the addressed problem (see caption for details).

\definecolor{color0}{rgb}{0.252663,0.332837,0.783665}
\definecolor{color1}{rgb}{0.275827,0.366717,0.812553}
\definecolor{color2}{rgb}{0.299441,0.400248,0.839842}
\definecolor{color3}{rgb}{0.323718,0.433158,0.864722}
\definecolor{color4}{rgb}{0.348323,0.465711,0.888346}
\definecolor{color5}{rgb}{0.373552,0.497499,0.909467}
\definecolor{color6}{rgb}{0.399231,0.528528,0.928459}
\definecolor{color7}{rgb}{0.425199,0.559058,0.946061}
\definecolor{color8}{rgb}{0.451739,0.588181,0.960201}
\definecolor{color9}{rgb}{0.478462,0.616564,0.972721}
\definecolor{color10}{rgb}{0.505423,0.643995,0.983157}
\definecolor{color11}{rgb}{0.532568,0.669801,0.990393}
\definecolor{color12}{rgb}{0.559747,0.694768,0.996075}
\definecolor{color13}{rgb}{0.586921,0.718121,0.998874}
\definecolor{color14}{rgb}{0.613933,0.739923,0.999142}
\definecolor{color15}{rgb}{0.640828,0.760752,0.997846}
\definecolor{color16}{rgb}{0.667253,0.779176,0.992959}
\definecolor{color17}{rgb}{0.693321,0.796314,0.986308}
\definecolor{color18}{rgb}{0.718985,0.811993,0.977656}
\definecolor{color19}{rgb}{0.743754,0.825125,0.965798}
\definecolor{color20}{rgb}{0.768034,0.837035,0.952488}
\definecolor{color21}{rgb}{0.791392,0.846750,0.936641}
\definecolor{color22}{rgb}{0.813693,0.854282,0.918480}
\definecolor{color23}{rgb}{0.835345,0.860514,0.898970}
\definecolor{color24}{rgb}{0.855378,0.863778,0.876587}
\definecolor{color25}{rgb}{0.875557,0.860242,0.851430}
\definecolor{color26}{rgb}{0.895882,0.849906,0.823499}
\definecolor{color27}{rgb}{0.912765,0.836682,0.794512}
\definecolor{color28}{rgb}{0.928116,0.822197,0.765141}
\definecolor{color29}{rgb}{0.940879,0.805596,0.735167}
\definecolor{color30}{rgb}{0.950956,0.786875,0.704761}
\definecolor{color31}{rgb}{0.959518,0.766973,0.674145}
\definecolor{color32}{rgb}{0.964835,0.744614,0.643239}
\definecolor{color33}{rgb}{0.968203,0.720844,0.612293}
\definecolor{color34}{rgb}{0.969851,0.695830,0.581312}
\definecolor{color35}{rgb}{0.968105,0.668475,0.550486}
\definecolor{color36}{rgb}{0.964911,0.640159,0.519806}
\definecolor{color37}{rgb}{0.959385,0.610306,0.489382}
\definecolor{color38}{rgb}{0.951254,0.578799,0.459408}
\definecolor{color39}{rgb}{0.941728,0.546413,0.429707}
\definecolor{color40}{rgb}{0.929357,0.512254,0.400673}
\definecolor{color41}{rgb}{0.915157,0.476927,0.372179}
\definecolor{color42}{rgb}{0.899534,0.440692,0.344107}
\definecolor{color43}{rgb}{0.880896,0.402331,0.317115}
\definecolor{color44}{rgb}{0.861054,0.362916,0.290628}
\definecolor{color45}{rgb}{0.839365,0.321856,0.264924}
\definecolor{color46}{rgb}{0.815508,0.277781,0.240294}
\definecolor{color47}{rgb}{0.790562,0.231397,0.216242}
\definecolor{color48}{rgb}{0.763520,0.178667,0.193396}
\definecolor{color49}{rgb}{0.735077,0.104460,0.171492}

\setlength{\fboxsep}{0mm}
\setlength{\tabcolsep}{2pt}
\renewcommand{\arraystretch}{1}

\begin{table*}[ht!]{

			\renewcommand{\tablename}{Figure}
			 \caption{\label{tab:svm_1} 
			Illustration of our text based clip retrieval approach for the action ``put wheel on a car''.
			We display the narration of the video that is obtained from Automatic Speech Recognition (ASR).
			The text is highlighted with different colors.
			These colors correspond to the score of the SVM that has been trained to detect sentences which refer to the action of interest.
			More precisely, for each word, we compute the average score over all the windows that contain it.
			Red indicates high score, blue indicates low score.
			The shown frames correspond to the top scoring part of the narration.
			We can see the coherence between what the person says (highlighted in red) and does in the video.
			Note the different challenges of the problem. 
			First, the input narration is very long.
			Second, the text directly comes from ASR, therefore it contains mistakes and does not have any punctuation.
			Third, several different expressions are similar and could refer to the action of interest.
			Despite all these challenges, our method is able to correctly retrieve the clip that contains the action of interest.}}
\end{table*}
\addtocounter{figure}{+1}

\section{Dataset of manipulated objects}
\label{app:dataset}

Table~\ref{tab:datasetstatistics} provides statistics for the dataset introduced in Section~\ref{sec:dataset} of the main paper.
For each object class we indicate associated action classes and the number of video clips for each action.
We also provide the list of states and the number of object tracklets with state annotations.
In total, we have around 20,000 annotated tracks which we use for the quantitative evaluation of state discovery.

\begin{table}[h!]
\centering
\resizebox{\columnwidth}{!}{
\begin{tabular}{@{}llll@{}}
\toprule
Objects    & \textbf{Actions} (\#clips)                     & \textit{States}                & \#Tracklets \\ \midrule
wheel      & \{\textbf{remove} (47),  \textbf{put} (46)\} & \textit{\{attached, detached\}} & 5447        \\
coffee cup & \{\textbf{fill} (57)\}                         & \textit{\{full, empty\}}        &  1819           \\
flower pot & \{\textbf{put plant} (27)\}                    & \textit{\{full, empty\}}        & 2463        \\
fridge     & \{\textbf{open} (234), \textbf{close} (191)\}   & \textit{\{open, closed\}}      & 7968        \\
oyster     & \{\textbf{open} (28)\}                              & \textit{\{open, closed\}}       & 1802        \\ \bottomrule
\end{tabular}
}
\caption{Statistics of our new dataset of manipulated objects \label{tab:datasetstatistics}}
\end{table}

  \begin{figure*}[t!]
  \begin{subfigure}{.44\textwidth}
    \raggedleft
    \includegraphics[width=\linewidth]{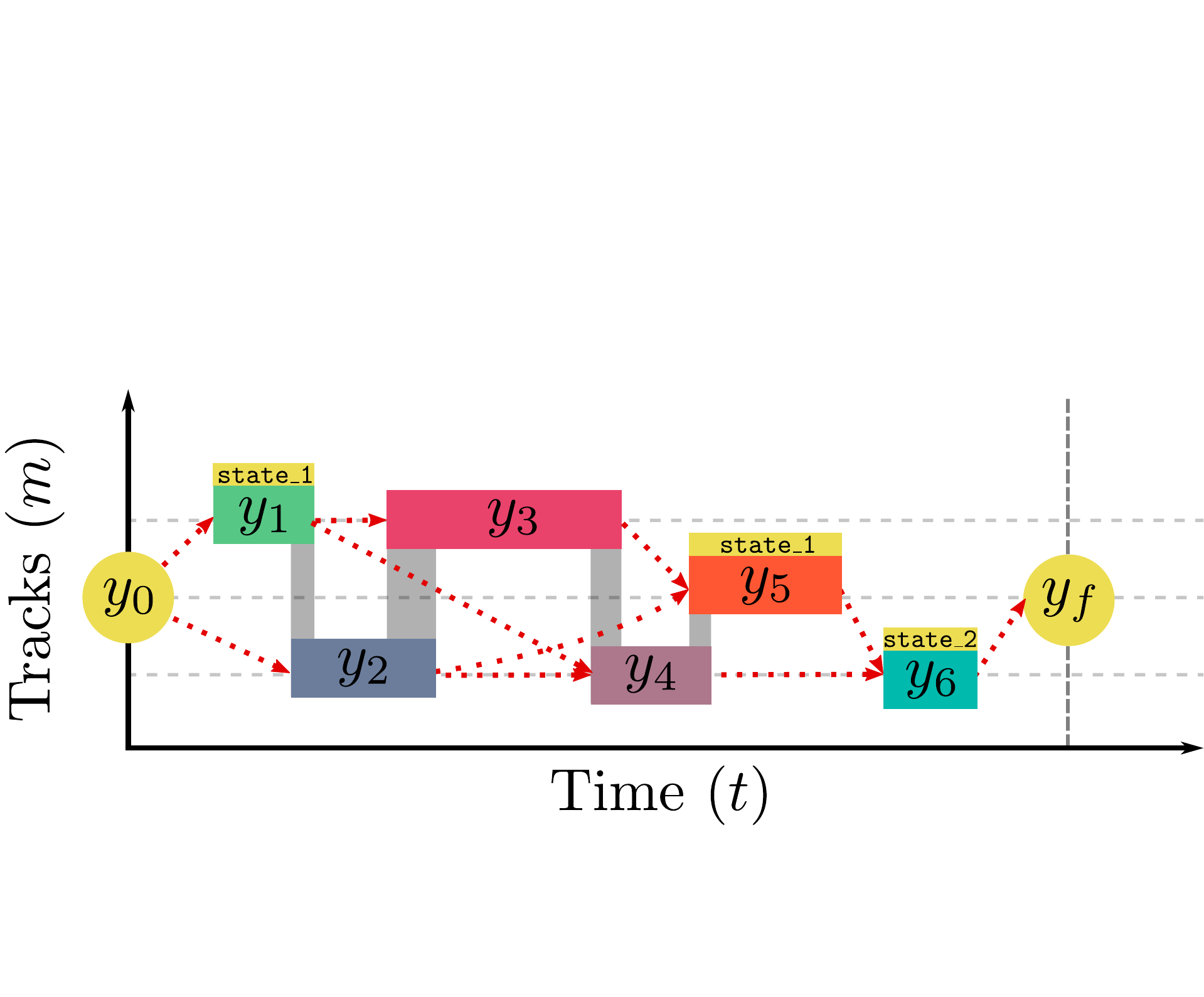}
    \caption{Non-overlap data structure for tracklets}
    \label{fig:DPsub1}
  \end{subfigure}
  \hspace*{\fill}
  \begin{subfigure}{0.53\textwidth}
    \raggedright
    \includegraphics[width=\linewidth]{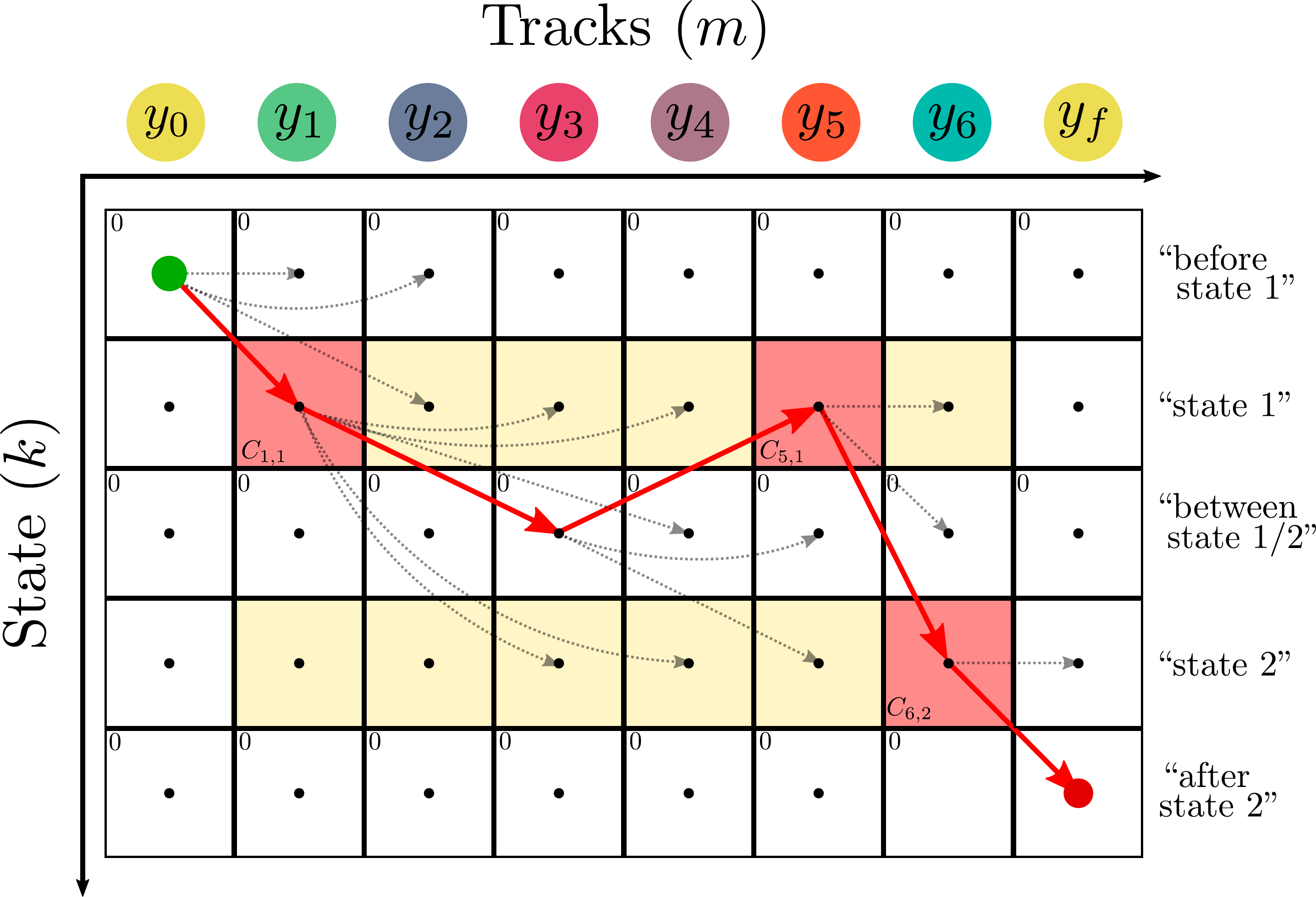}
    \caption{Cost matrix $\tilde{C}_n$ for the dynamic program}
    \label{fig:DPsub2}
  \end{subfigure}
  \caption{\small
In \textbf{(a)}, we provide an illustration of a possible situation for the tracklets.
 $y_0$ and $y_f$ are two fictitious tracklets that encode the beginning and end of the video.
 Each tracklet is indexed based on its beginning time.
 The time overlap between tracklets is shown by the grey color.
 We specify for each tracklet its possible successors by the dotted red arrows (see main text).
 Finally an admissible labeling is illustrated by yellow tags where $y_1$ and $y_5$ have both been assigned to state~$1$ and $y_6$ to state~$2$.
In \textbf{(b)}, we give an illustration of our approach to solve~\eqref{eq:linearprogram} with a dynamic program.
We display the modified cost matrix $\tilde{C}_n$ (see main text).
A valid path has to go from the green dot ($y_0$) to the red dot ($y_f$).
The light yellow entries show part of the $C_n$ matrix that are inserted in $\tilde{C}_n$, whereas white entries encode the rows of $0$s that are inserted to impose the \textbf{at least one} ordering constraint.
The red arrows specify an example optimal path inside the matrix.
The red entries display the tracklets that have been assigned to state~$1$ ($y_1$ and $y_5$) or state $2$ ($y_6$) (equivalent to putting ones in the appropriate corresponding entries in $Y_n$).
Finally, the grey arrows display the possible valid transitions that can be made for the entries \emph{along the red path}, for clarity. We see for example that from $(2,y_1)$, there are $6$ possible transitions: two column choices from the two red arrows from $y_1$ in~(a) encoding the non-overlap constraint; and three row choices encoding the valid transition from ``state 1'' (corresponding to the choice ``state 1'', $0$ or ``state 2'' for the next tracklet) encoding the ``at least one'' ordering constraint.}
  \label{fig:DP}
\end{figure*}

\section{Dynamic program for the tracklets}
\label{app:dp}
The track constraints defined in Section~\ref{sec:statemodel} introduce new challenges compared to the previous related work~\cite{Alayrac15Unsupervised,Bojanowski14weakly,Joulin14efficient}.
Recall that there are three main components in the constraints.
First, we assume that only one object is manipulated at a given time. 
Thus \emph{at most one} tracklet can be assigned to a state at a given time.
This constraint is referred to as the \textbf{non-overlap constraint}.
Second, we have the \textbf{ordering constraint} that imposes that \emph{state~1} always happens \emph{before} \emph{state~2}.
The last constraint imposes that we have \textbf{at least one} tracklet labeled as \emph{state~1} and \emph{at least one} tracklet labeled as \emph{state~2}.
We need to be able to minimize linear functions over this set of constraints in order to use the Frank-Wolfe algorithm.
More precisely, as the constraints decompose over the different clips, we can solve independently for each clip $n$ the following linear problem:
\begin{align}
\label{eq:linearprogram}
\underset{\substack{Y_n\in\{0,1\}^{M_n\times 2}}}{\text{minimize}}  & \phantom{aa:}\text{Tr}(C_n^TY_n) \\
\text{ s.t. }   & \underbrace{Y_n \in \mathcal{Y}_n}_{\substack{\text{non-overlap + ordering}\\\text{+ at least one}}},   \nonumber
\end{align}
where $C_n\in\mathbb{R}^{M_n\times 2}$ is a cost matrix that typically comes from the computation of the gradient of the cost function at the current iterate.
In order to solve this problem, we use a dynamic program approach that we explain next.
Recall that we are given $M_n$ tracklets $(y_i)_{i=1}^{M_n}$ and our goal is to output the $Y_n$ matrix that assigns to each of these tracklets either state~$1$, state~$2$ or no state at all while respecting the constraints.
The whole method is illustrated in Figure~\ref{fig:DP} with a toy example.

\textbf{Non-overlap data structure for the tracklets.}
We first pre-process the tracklets to build an auxiliary data-structure that is used to enforce the non-overlap constraint between the tracklets, as illustrated in Figure~\ref{fig:DPsub1}.
First, we sort and index each tracklet by their beginning time, and add two fictitious tracklets: $y_0$ as the starting tracklet and $y_f$ as the ending tracklet.
These two tracklets are used to start and terminate the dynamic program.
If all the tracklets were sequentially ordered without any overlap in time, then we could simply make a decision for each of them sequentially as was done in previous work on action localization for example (one decision per time step)~\cite{Bojanowski14weakly}.
To enforce the non-overlap constraint, we force the decision process to choose \emph{only one} possible successor among the group of overlapping valid immediate successors of a tracklet. For each tracklet $y_i$, we thus define its (smallest) set of ``\emph{valid successors}'' as the earliest tracklet $y_j$\footnote{Earliest means the smallest $j$.} after $y_i$ that is also non-overlapping with $y_i$, as well as any other tracklet $y_l$ for $l > j$ that is overlapping with $y_j$ (thus giving the earliest valid group of overlapping tracklets). The valid successors are illustrated by red dotted arrows in Figure~\ref{fig:DPsub1}. For example, the valid successors of~$y_1$ are $y_3$ (the earliest one that is non-overlapping) as well as $y_4$ (which overlaps with $y_3$ thus forming an overlapping group). Skipping a tracklet in this decision process means that we assign it to zero (which trivially always satisfies the non-overlapping constraint); whereas once we choose a tracklet to potentially assign it to state 1 or 2, we cannot visit any overlapping tracklet by construction of the valid successors, thus maintaining the non-overlap constraint.

\textbf{Dynamic program.}
The dynamic programming approach is used when we can solve a large problem by solving a sequence of inclusive subproblems that are linked by a simple recursive formula and that use overlapping solutions (which can be stored in a table for efficiency).
In terms of implementation, \cite{Bojanowski14weakly} encoded their dynamic program as finding an optimal path inside a cost matrix.
This approach is particularly suited when the update cost rule depends \emph{only} on the arrival entry in the cost matrix as opposed to be transition dependent.
As we will show below, we can encode the solution to our problem in a way that satisfies this property.
We therefore use the framework of~\cite{Bojanowski14weakly} by casting our problem as a search for an optimal path inside a cost matrix $\tilde{C}_n$ illustrated in Figure~\ref{fig:DPsub2}, and where the valid transitions encode the possible constraints.

One main difference with~\cite{Bojanowski14weakly} is that we have to deal with the challenging \textbf{at least one} constraint in the context of ordered labels.
To do so, we can filter further the set of valid decisions by using ``memory states'' that encode in which of the following three situations we are: \textbf{(i)} that state~$1$ has not yet been visited, \textbf{(ii)} that state~$1$ has already been visited, but state~$2$ has not yet been visited (and thus that we can either come back to state~$1$ or go to state~$2$) and \textbf{(iii)} that both states have been visited.
These memory states can be encoded by interleaving complete rows of $0$s in between columns of $C_n$ stored as rows, to obtain the $5 \times M_n$ matrix~$\tilde{C}_n$.
These new rows encode the three different memory states previously described when making a prediction of $0$ for a specific tracklet, and we enforce the correct memory semantic by only allowing a path to move to the same row or the row immediately below, except for state~$1$ which can also move directly to state~$2$ (two rows below), and the middle ``between state 1/2'' row, where one can go up one row additionally to state~$1$. Finally, the valid transitions between columns (tracklets) are given by the \emph{valid successors} data structure as given in Figure~\ref{fig:DPsub1} to encode the non-overlap constraints. Combining these two constraints (at least one ordering and non-overlap), we illustrate with grey arrows in Figure~\ref{fig:DPsub2} the possible transitions from the states along the path in red. To describe the dynamic program recursion below, we need to go the opposite direction from the successors, and thus we say that $y_j$ is a \emph{predecessor} of $y_i$ if and only if $y_i$ is a successor of $y_j$.

To perform the dynamic program, we maintain a matrix $D_n$ of the same size as $\tilde{C}_n$ where
$D_n(k,i)$ contains the minimal valid path cost of going from $(1,y_0)$ to $(k,y_i)$ inside the cost matrix $\tilde{C}_n$.
To define the cost update recursion to compute $D_n(k,i)$, let $P(k,i)$ be the set of tuples $(l,j)$ for which it is possible to go from $(l,j)$ to $(k,y_i)$ according to the rules described above.  
The update rule is then as follows:
\begin{align}
\label{eq:updateruleDP}
D_n(k,i) = \min_{(l,j) \in P(k,i)} D_n(l,j)+\tilde{C}_n(k,y_i).
\end{align}
As we see here, the added cost depends \emph{only} on the arrival entry $\tilde{C}_n(k,y_i)$.
We can therefore use the approach of~\cite{Bojanowski14weakly} and only consider entry costs rather than edge costs.
Thanks to our indexing property (tracklets are sorted by the beginning time), we can update the dynamic program matrix by filling each column of $D_n$ one after the other.
Once this update is finished, we back-track to get the best path by starting from the ending track (predecessors of $y_f$) at the last row (to be sure that both states have been visited) that has the lowest score in the $D_n$ matrix.
The total complexity of this algorithm is of order $\mathcal{O}(M_n)$.

\section{Joint cost rounding method}
\label{app:jcr}

Recall that we propose to use a convex relaxation approach in order to obtain a candidate solution of main problem~\eqref{eq:jointstateactionprob}.
Thus, we need to \emph{round} the relaxed solution afterward in order to get a valid integer solution.
We propose here a new rounding that is adapted to our joint problem.
We referred to this rounding as the \textbf{joint cost rounding} (see Section~\ref{sec:optim} of main paper).
This rounding is inspired by~\cite{Bojanowski15weakly,Alayrac15Unsupervised}.
They observe that using the learned $W^*$ classifier to round gives them better solutions, both in terms of objective value and performance.
We propose to use its natural extension for our joint model.
We first fix the classifiers for actions $W_a$ and for states $W_s$ to their relaxed solution $(W_a^*, W_s^*)$ and find, for each clip $n$, the couple $(Z_n,Y_n)$ that minimizes the joint cost~\eqref{eq:jcrrounding}.
To do so, we observe that we can enumerate all $T_n$ possibilities for $Z_n$, and solve for each of them the minimization of the joint cost with respect to $Y_n$.
The minimization with respect to $Y_n$ can be addressed as follows.
First, we observe that the distortion function~\eqref{eq:relaxeddist} is bilinear in $(Z_n,Y_n)$.
Let $Z_n$ be a $T_n \times 1$ vector, and let $\mathbf{1}_2$ be a vector of ones of length 2.
We can actually write: $d(Z_n,Y_n)=\text{Tr}(( B_n Z_n \mathbf{1}_2^\top)^\top Y_n)$ for some matrix $B_n\in\mathbb{R}^{M_n \times T_n}$.
Thus, when $Z_n$ is fixed, the joint term $d(Z_n,Y_n)$ is actually a simple \emph{linear} function of $Y_n$.
In addition, the quadratic term in $Y_n$ coming from~\eqref{eq:discr_state} is also \emph{linear} over the integer points (using the fact that $y^2=y$ for $y\in\{0,1\}$). 
Thus, when $Z_n$, $W_a^*$ and $W_s^*$ are fixed, the minimization over $Y_n$ is a linear program~\eqref{eq:linearprogram} that we solve using our dynamic program from the previous section.
The final algorithm is given in Algorithm~\ref{jcrAlgo}.
Its complexity is of order $\mathcal{O}(T_nM_n)$.

\definecolor{comment}{RGB}{86, 115, 154}
\begin{algorithm}[t!]
	\caption{Joint cost rounding for video $n$}
	\label{jcrAlgo}
	\begin{algorithmic}
		\State Get  $W_s^*$ and $W_a^*$ from the relaxed problem.
		\State Initialize $Z^*$, $Y^*$ and $\text{val}^*=+\infty$.
		\State {\color{comment} \footnotesize \# Loop over all possibilities for $Z_n$ (saliency)}
		\For{ $t$ in 1 : $T_n$}
			\State $Z$ $\gets$ \texttt{zeros}($T_n$, $1$) {\color{comment} \footnotesize \# Set the $t$-th entry of $Z$ to $1$}
			\State $Z_t$ $\gets$ $1$ 			
			\State {\color{comment} \footnotesize \# Definition of the cost matrix}
			\State $C_n$ $\gets$ $\frac{1}{2M}\left(\texttt{ones}(M_n, 2)-2X_sW_s\right)+\frac{\nu}{T} B_n Z \mathbf{1}_2^\top$ 
			\State {\color{comment} \footnotesize \# Dynamic program for the tracks}
			\State $Y_{\min}$ $\gets$ $\argmin_{Y \in \mathcal{Y}_n} \text{Tr}(C_n^TY)$ 
			\State {\color{comment} \footnotesize \# Cost computation}
			\State $\text{cost}_Z \gets \frac{1}{2T} \|Z - X_{a} W_{a}\|_F^2$
			\State $\text{cost}_Y \gets \frac{1}{2M} \|Y_{\min} - X_{s} W_{s}\|_F^2$
			\State $\text{cost}_{ZY} \gets \frac{\nu}{T}d(Z,Y_{\min})$
			\State {\color{comment} \footnotesize \# Update solution if better}
			\State $\text{val} \gets \text{cost}_Z + \text{cost}_Y + \text{cost}_{ZY}$
			\If{$\text{val} < \text{val}^*$} 
				\State {$Z^*$ $\gets$ $Z$} 
				\State {$Y^*$ $\gets$ $Y_{\min}$} 
				\State {$\text{val}^*$ $\gets$ $\text{val}$} 
			\EndIf
		\EndFor
		\State\Return $Z^*,Y^*$
	\end{algorithmic}
\end{algorithm}

\section{Supervised baselines for Action Localization}
\label{app:supervised}

\begin{table}[t!]
	\centering
	\resizebox{\linewidth}{!}{
		\begin{tabular}{@{}cccccccc|c@{}}
			\toprule
			\multirow{2}{*}{\textbf{Features}}   & put                       & remove                    & fill                    & open                       & fill                         & open                       & close                       & \multirow{2}{*}{\textbf{Average}} \\
			& \multicolumn{1}{l}{wheel} & \multicolumn{1}{l}{wheel} & \multicolumn{1}{l}{pot} & \multicolumn{1}{l}{oyster} & \multicolumn{1}{l}{coff.cup} & \multicolumn{1}{l}{fridge} & \multicolumn{1}{l|}{fridge} &                                   \\ \midrule
			\multicolumn{1}{l}{\textbf{(1)} CNN + HOF} & \textbf{0.65}                      & 0.68                      & 0.56                    & 0.11                       & 0.91                         & 0.54                       & 0.59                        & 0.58                              \\
			\multicolumn{1}{l}{\textbf{(2)} CNN + IDT}        & \textbf{0.65}                      & \textbf{0.72}                      & \textbf{0.56}                    & \textbf{0.21}                       & \textbf{0.93}                         & \textbf{0.6}                        & \textbf{0.62}                        & \textbf{0.61}                              \\ \bottomrule
		\end{tabular}
	}
	\caption{ \small \label{tab:supexp} Results of supervised baselines for action localization.}
\end{table}

We have run supervised baseline methods with state-of-the-art features. 
To be able to compare numbers with our experiment, we used a leave-one-out technique. 
For each action, we train a binary classifier with SVM on all videos except one. 
Similarly to our setting, we then select the top scoring time interval of the left alone test video. 
We repeat this process for all videos and report the metric used in our paper. 
For baseline \textbf{(1)}, we use the same features we are using in the main paper.
For baseline \textbf{(2)}, we complete our features with all channels of Improved Dense Trajectories (IDT)~\cite{Wang13action}.
Detailed results are given in Table~\ref{tab:supexp}.
We observe that we obtain results that are on par with our weakly supervised baselines (0.55 versus 0.58), therefore demonstrating the potential of using the information of object states for action localization.

\section{Failure cases}
\label{app:failcases}

\begin{figure}[t!]
	\centering
	\includegraphics[width=\linewidth]{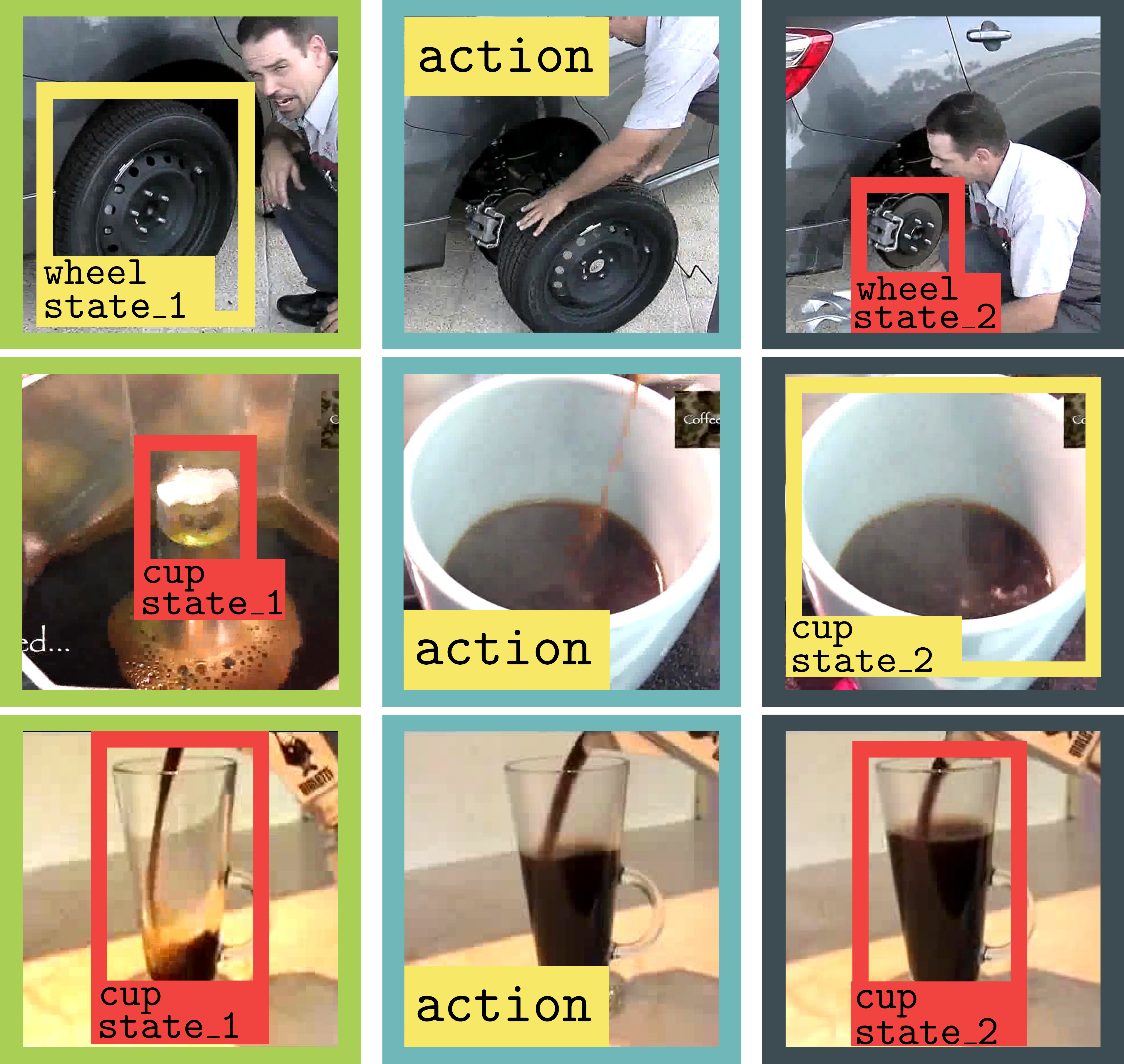}
	\caption{
		Typical failure cases for ``\textit{removing car wheel}'' (top) and
		`̀`\textit{fill coffee cup}'' (middle, bottom) actions. 
		Yellow indicates correct predictions; red indicates mistakes. Top: the removed wheel is incorrectly localized (right). 
		Middle: the ``empty cup'' is incorrectly
		localized (left). 
		Bottom: In this case, both object tracklets are annotated as ``\textit{ambiguous}'' in the ground truth as they occur during
		the pouring action and hence the predictions, while they appear
		reasonable, are deemed incorrect.
	}
	\vspace{-2mm} %
	\label{fig:failcases}
\end{figure}	

We observed two main types of failures, illustrated in Figure~\ref{fig:failcases}. 
The first one occurs when a false positive object detection consistently satisfies the hypothesis of our model in multiple videos (the top two rows in Figure~\ref{fig:failcases}).
The second typical failure mode is due to ambiguous labels (bottom row in Figure~\ref{fig:failcases}). 
This highlights the difficulty in
annotating ground truth for long actions such as ``\textit{pouring coffee}".

\end{document}